\documentclass[runningheads]{llncs}

\usepackage{graphicx}
\usepackage{comment}
\usepackage{amsmath,amssymb}
\usepackage{color}
\usepackage{url}
\usepackage{hyperref}
\usepackage[nottoc]{tocbibind}
\usepackage[square,numbers]{natbib}
\newif\ifreview
\reviewfalse

\ifreview
	\usepackage{lineno}

	\linenumbers
\fi

\begin{document}


\def\SubNumber{000}

\def\GCPRTrack{Regular Track}

\title{Show Me Your Face, And I’ll Tell You How You Speak}

\ifreview
	\titlerunning{DAGM GCPR 2021 Submission \SubNumber{}. CONFIDENTIAL REVIEW COPY.}
	\authorrunning{DAGM GCPR 2021 Submission \SubNumber{}. CONFIDENTIAL REVIEW COPY.}
	\author{DAGM GCPR 2021 - \GCPRTrack{}}
	\institute{Paper ID \SubNumber}
\else
    
    
	\author{Christen Millerdurai\inst{1}\thanks{All authors contributed equally(Listed alphabetically).} \and Lotfy Abdel Khaliq\inst{2}$^*$  \and Timon Ulrich\inst{3}$^*$ }

	\authorrunning{ }
	
	\institute{Saarland University}
	
\fi

\maketitle              

\begin{abstract}

When we speak, the prosody and content of the speech can be inferred from the movement of our lips. In this work, we explore the
task of lip to speech synthesis, i.e., learning to generate speech given only the lip movements of a speaker where we focus on learning accurate
lip to speech mappings for multiple speakers
in unconstrained, large vocabulary settings. We capture the speaker's voice identity through their facial characteristics, i.e., age, gender, ethnicity and condition them along with the lip movements to generate speaker identity aware speech. To this end, we present a novel method "Lip2Speech",  with key design choices to achieve accurate lip to speech synthesis in unconstrained scenarios. We also perform various experiments and extensive evaluation using quantitative, qualitative metrics and human evaluation.

\end{abstract}
\begin{figure}[htb]
\includegraphics[width=\textwidth]{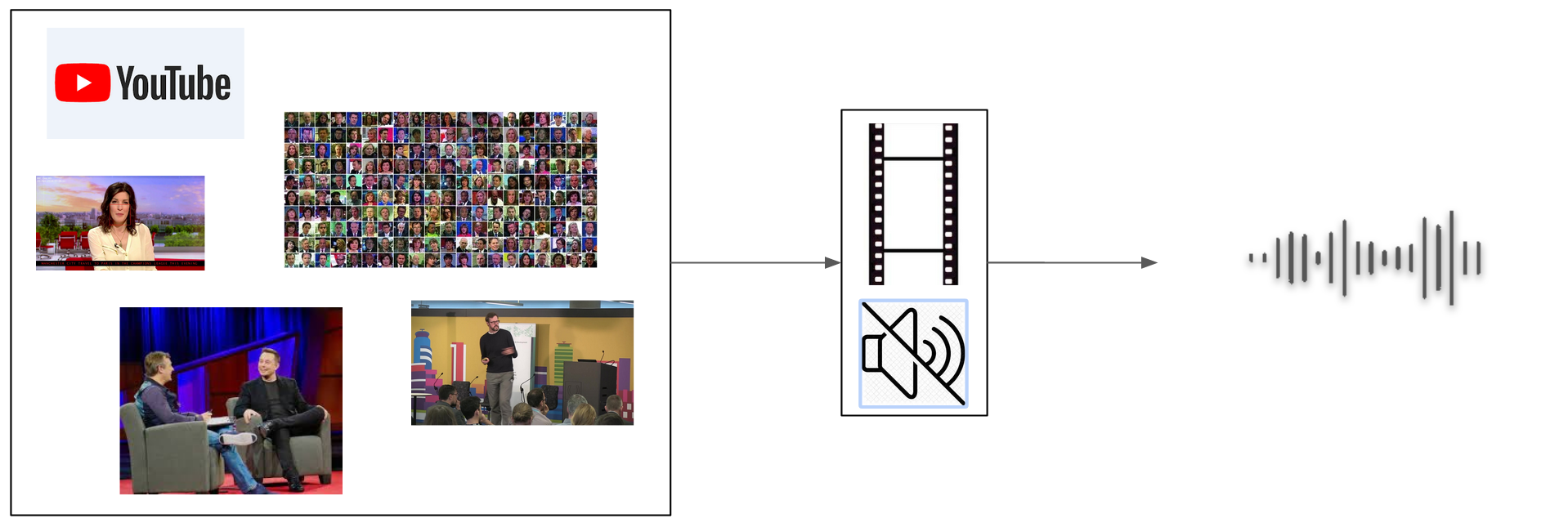}
\caption{Our objective is taking a silent video of a person speaking and generate speech from their lip movements.} 
\label{fig:motiv}
\end{figure}

\section{Introduction}

Lip reading is a technique of understanding speech by visually interpreting the movements of the lips. Although lip reading is used most extensively by deaf and hard-of-hearing people, most people with normal hearing process some speech information from sight of the moving mouth \cite{Woodhouse2009}. Also, understanding linguistic cues from readings the lips can improve the clarity of the conversation in environments with noisy backgrounds. One possible approach for lip reading is to generate text for the corresponding lip movements. Although this method works, humans also emote through their lip movements which is lost in the textual representation. Also, they cannot represent non-linguistic vocal repertoire\cite{Anikin2017}. Moreover, several use-cases like voice inpainting\cite{zhou2019visioninfused}, speech recovery from background noise\cite{Ephrat_2018}, generating a voice for people who cannot produce voiced sounds (aphonia) are not possible. While Lip-to-text-to-speech might solve certain problems, the intermediate textual information distills the prosody, tempo, non-verbal cues of the lip movements which are essential to solve most of the problems mentioned above.Hence, both Lip-to-text, Lip-to-text-to-speech are at a disadvantage and a direct lip-to-speech method is needed. Moreover, text-based systems require transcribed datasets for training, which are hard to obtain because they require laborious manual annotation. And training of Lip-to-speech methods can be done in a self-supervised manner since most video comes paired with the corresponding audio. While, we have established Lip-to-speech methods are superior to compared to other methods, generating voice requires different vocal identity for each person to bolster realism. Hence, the Lip-to-speech methods should be conditioned on the speaker's vocal identity while generating the speech content. While methods like \cite{Gibiansky2017DeepV2, Nachmani2018FittingNS, Arik2018NeuralVC, Ping2017DeepV3, jia2019transfer} sample the voice from the target speaker which are then conditioned to generate the content, we assume that a voice to be "cloned" is not available at test time. Hence, we rely on facial features such as the speaker's gender, age, ethnicity to generate a voice which is then conditioned along with the lip movements to generate the speech.

While several methods\cite{mira2021endtoend,vougioukas2019videodriven,um2021facetron} have tried to solve lip to speech synthesis, most methods focus on clinical settings, i.e GRID corpus\cite{Alghamdi2018} but our work  mainly focuses on lip reading in the wild. We draw parallel to two works, Lip2Wav\cite{prajwal2020learning} and Facetron\cite{um2021facetron}, while Lip2Wav also accomplishes our objective, its multi-speaker pipeline is conditioned on the speaker's voice, which if not available will not work. While Facetron does not have this issue, its GAN\cite{goodfellow2014generative} based decoder limits its capability to generate very long sequences. Moreover, even though Facetron uses the face to condition the speaker's voice, it is not demonstrated on Lip reading in the wild(LRW)\cite{Chung16} dataset. 
In this work, we propose "Lip2Speech", an end-to-end model that is capable of directly converting silent video to melspectogram, which can converted to raw audio either by Griffin-Lim Algorithm\cite{1172092} or neural vocoders\cite{oord2017parallel,oord2016wavenet}.

Our method is based on encoder-decoder architecture, where we encode the lip movements along with the face features and decode the melspectorgram of the generated speech. The generated speech is evaluated using standard audio reconstruction and speech synthesis metrics. Additionally, we propose YouTube Lip Data(YLD), a pipeline that can create training data from any YouTube video. 

We release the code, trained models publicly for future research along with a demonstration video here\footnote{ \url{https://github.com/Chris10M/Lip2Speech}}. The
rest of the paper is organized as follows: In Section 2, we
survey the recent developments in this space. Following
this, we describe our novel Lip2Speech network in Section 3. Our datasets, training details and implementation details are explained in Section 4. And in Section 5, we perform various experiments with our method and introduce the YLD pipeline. And finally, we conclude our work in Section 8.

\section{Related work}
\subsection{Face reconstruction from Audio}
Several methods have been proposed to infer visual information from  speech signals. There are two research directions for this problem: one more towards graphic-based generation\cite{10.1145/3072959.3073640,10.1145/3072959.3073699} and the other towards pixel-level generation. However, graphic-based methods typically parametrize the reconstructed face using a priori face model and texture. In pixel-level generation, Sadoughi et al.\cite{sadoughi2018speechdriven} use speech to reconstruct lip motions. Duarte et al.\cite{duarte2019wav2pix} reconstruct the face including the expression and pose from speech using GANs. While several methods have explored inferring separate face traits\cite{8316819,inproceedings}, William et al\cite{oh2019speech2face}. did not restrict the model to learn pre-defined facial traits but rather a more general visual representation of how the person speaking looks like.

\subsection{Lip reading}

Many end-to-end approaches have been proposed to tackle this problem and can be divided into two categories. In the first category, an MLP is used to extract features and LSTM layers model the temporal dynamics of the sequence\cite{f2ec7e83b30342a3983cea304b4b8732,wand2016lipreading}. In the second category, a 3D conv layer is used followed by standard conv layers combined with LSTMs. However, these methods do not extract features from audio directly, and they rely on MFCCs as a feature extractor and an attention mechanism is applied to lip crops and MFCCs.\cite{Chung_2017,doi:10.1177/1729881420976082}. Current state of the art methods combine all three, i.e.,  applies  spatio-temporal convolution to the lip ROIs followed by a spatial feature extractor coupled with LSTMs.  

\subsection{Lip to Text Generation}
Although multiple methods\cite{wand2016lipreading,Xu2018LCANetEL} were proposed to generate text from lip movements, most of them are limited to small vocabulary space. Also, there has been multiple approaches that tackle the problem in an in-the-wild manner by using sequence-to-sequence transformer\cite{NIPS2017_3f5ee243} models to generate sentences given a silent lip movements sequence.  

\subsection{Speech Synthesis}
Despite the fact that a lot of research has been investigated in generating natural speech from text, it still remains a challenging task. Existing statistical parametric speech synthesis\cite{4218329,6495700} use a vocoder to generate speech from trajectories of speech features. However, they produce an unnatural sounding voices. Recent works such as WaveNet\cite{oord2016wavenet} produces audio quality that is very close to the real human speech but the inputs to the model (linguistic features, predicted log fundamental frequency(F0), and phoneme duration) are hard to produce and requires significant domain expertise. Tacotron\cite{Wang2017TacotronTE}, a sequence-to-sequence architecture for producing melspectrograms from a sequence of characters, mitigates this issue by replacing the production of these linguistic features with a single neural network trained from data alone. Other approaches strive to generate speech with the voice of different speakers. Jia et al.\cite{jia2019transfer} proposed a model that is composed of three independently trained neural networks, (i) a speaker encoder\cite{wan2020generalized}, (ii) a sequence-to-sequence synthesizer\cite{shen2018natural} (iii) a neural vocoder\cite{oord2016wavenet}

\subsection{Lip to Speech Generation}

End-to-end methods as Vid2Speech\cite{7953127} and Lipper\cite{Kumar2019LipperST} extract LPC (Linear Predictive Coding) features for k subsequent frames followed by  applying a 2-D CNN. The drawback of such methods is that head motion is not taken into consideration so it does not model a real-world scenario. Another approach was taken to improved speech quality by generating waveforms using GANs but they do not make use of of sequence-to-sequence paradigm that is used for text-to-speech generation which can greatly improve the speech quality. Furthermore, these works provide results on a narrow vocabulary and very minimal head motion which might be ill-suited for real-world scenarios. Recent works such as \cite{prajwal2020learning} address these issues by 
using spatio-temporal encoder coupled with attention based sequence-to-sequence decoder for high-quality speech generation.

\begin{figure}[htb]
\includegraphics[width=\textwidth]{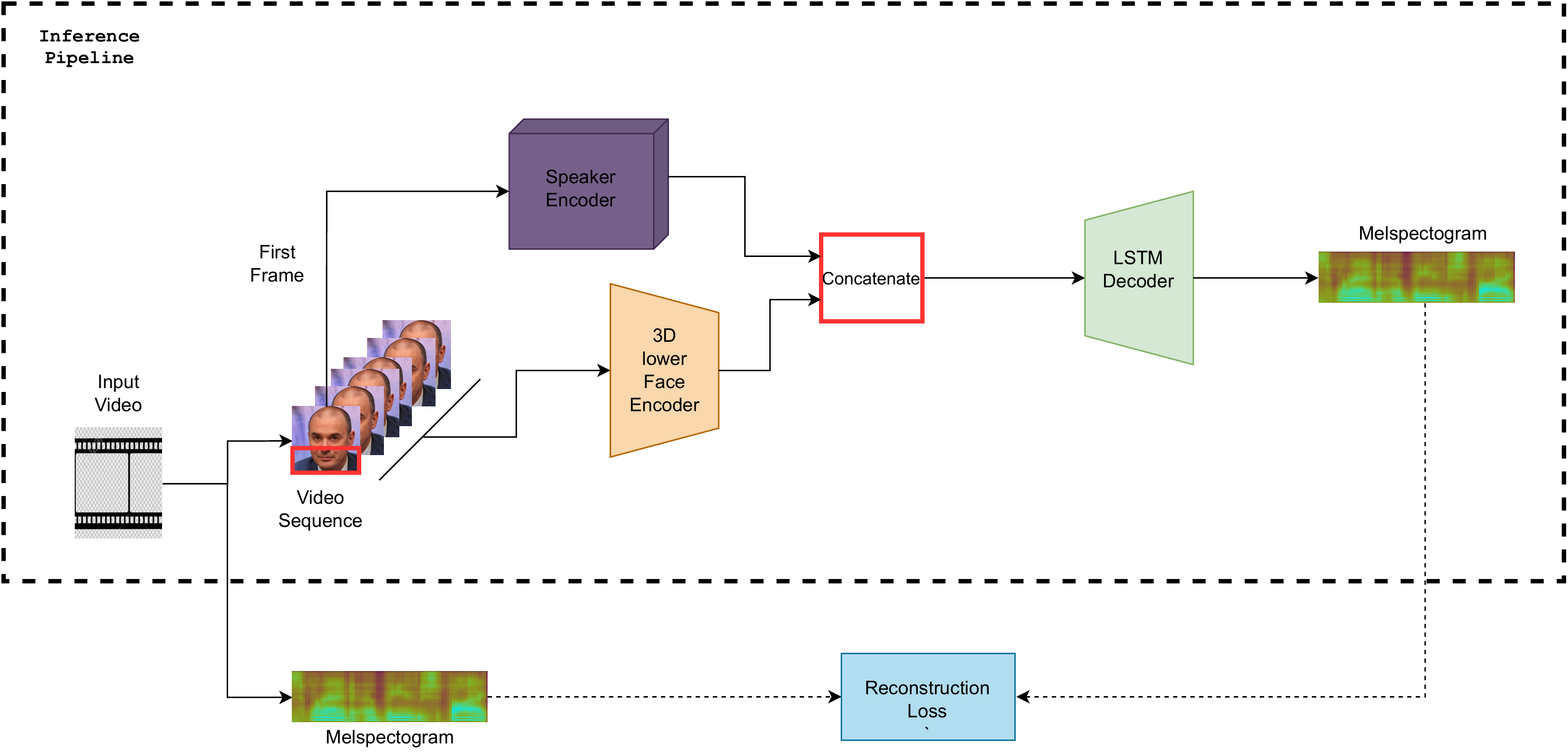}
\caption{The general overview of our architecture.} 
\label{fig:arch}
\end{figure}

\section{Lip2Speech Network}

In this section, we introduce our proposed network, Lip2Speech. We start by first introducing the problem statement, then explain the speaker identity module, followed by explaining the Lip-reading module and the speech synthesis module.

\subsection{Problem Statement}

Given an input video of a speaking face, our objective is to synthesize the speech of the speaking face with the identity of the speaker estimated through their face attributes, i.e., gender, age, ethnicity. The video contains a sequence of image frames, F = $(F_1, F_2, F_3, ....., F_N)$, and while the synthesized speech contains speech frames of the melspectogram, S = $(S_1, S_2, S_3, ....., S_T)$ . And this can be formulated as a sequence-to-sequence learning task\cite{sutskever2014sequence}, where the correspondences between the image frames and speech frames are learnt. These correspondences capture both the speaker identity and speech content. And, after learning the correspondences, speech frames can be synthesized for the image frames in an auto-regressive manner. It is to be noted that the number of image frames, $N$ and number of speech frames, $T$ does not need to be equal. Concretely, each output speech frame, $S_t$ is modelled as a conditional distribution of the previous speech frame and image frames, F.

$$
P(S|F) = \prod^{t=T}_{t=1}(S_t|S<t, F)
$$

\subsection{Architecture Overview}

For an input frame sequence F = $(F_1, F_2, F_3, ....., F_N)$, the first frame is passed to the speaker encoder to get the "speaker embeddings". Also for every frame $F_n$, the lip ROIs are cropped and passed to the 3D lower face encoder to get the face features. The speaker embeddings are tiled and then concatenated to the face features along the frame sequence $N$, which we call as "visual features". These visual features are passed to the LSTM decoder which generates the melspectogram of the speech in an auto-regressive manner. As shown in \autoref{fig:arch}, during training time we use the melspectogram from the input video and minimize the mean square error to the generated melspectogram.

\subsection{Speaker Identity Module}

\begin{figure}
\includegraphics[width=\textwidth]{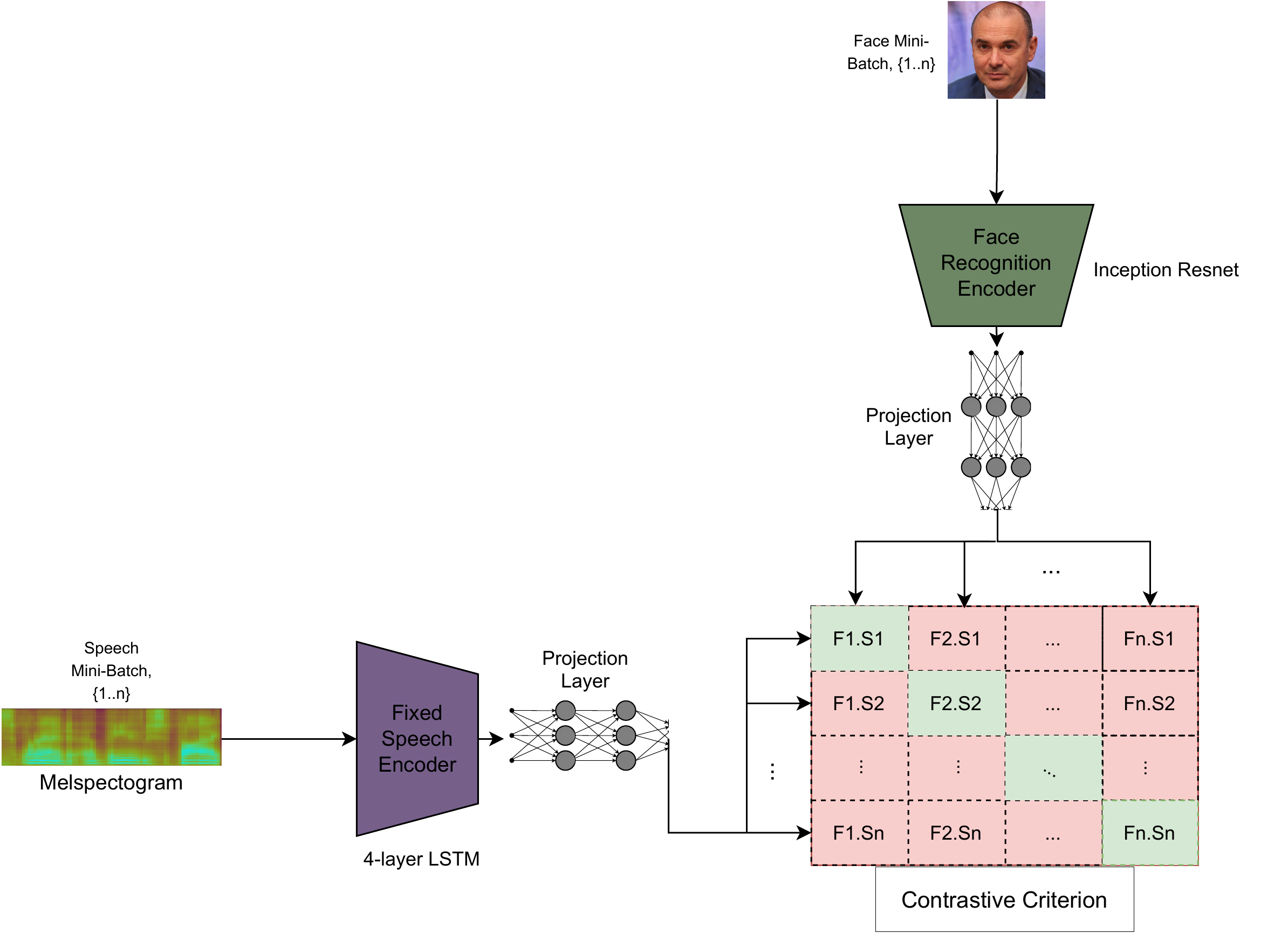}
\caption{The speaker encoder is trained to discriminate  } \label{fig:spekenc}
\end{figure}

The speaker identity module, as shown in \autoref{fig:spekenc} is used to learn the correlations between a speaker's face and their voice. We know from \cite{KAMACHI20031709, nagrani2018seeing}, humans have the ability to "put a face to a voice" and \cite{oh2019speech2face} tries to generate a face using a given voice sample. In our work, we try to "produce a voice to a face", an inverse of previous work. And we do this by generating encodings which are tied to every individual speaker. While \cite{Gibiansky2017DeepV2, Nachmani2018FittingNS, Arik2018NeuralVC, Ping2017DeepV3, jia2019transfer} also use encodings to map the speaker identity, they condition them using the speaker's voice. On the contrary, our work conditions the encodings from the face of the speaker. 

We use a Inception-ResNet\cite{szegedy2016inceptionv4} that is pre-trained on CASIA-WebFace\cite{yi2014learning} as the face recognition encoder. We freeze the first 3 blocks of the network and fine-tune the reset, i.e., 2 blocks. This network taken in a face and produces a face encoding. Our objective is to, learn the correspondences of the face that correlates to the speakers voice, hence we use a another network, speech encoder\cite{jia2019transfer}. The speech encoder takes in a melspectogram and generates a speaker encoding that correlates to the speaker's identity. And the speech encoder is frozen throughout the training.    

We learn a cross-modal mapping between the face encodings and the speaker encodings, which we define as "speaker embeddings" through instance based contrasting learning.

\subsection{Video feature extractor}

\begin{figure}
\includegraphics[width=\textwidth]{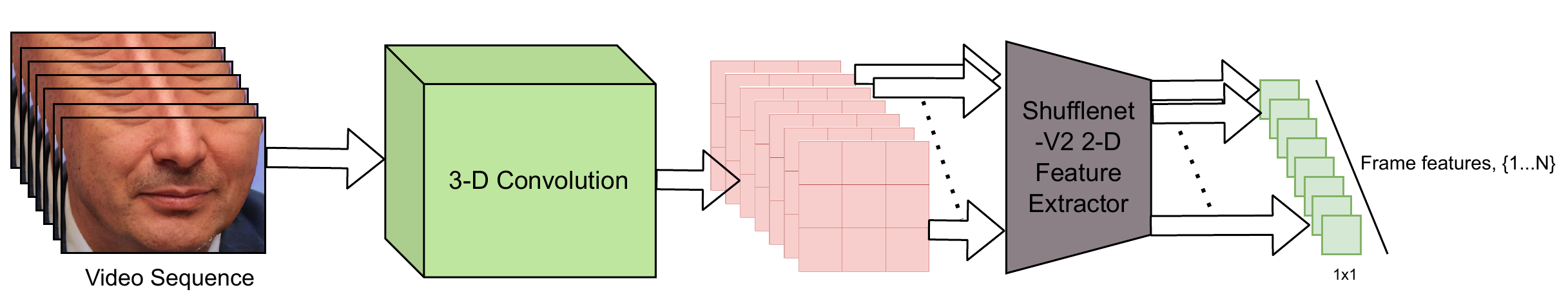}
\caption{The video feature extractor encodes the frame sequence by applying spatio-temporal convolutions.} \label{fig:enc}
\end{figure}

The Video feature extractor, as shown in \autoref{fig:enc} is used to extract the semantic and temporal lip features from the video frame sequence, F. We pass F = $(F_1, F_2, F_3, ....., F_N)$ to a 3-D convolution block to induce temporal awareness of the frame features and then apply a 2-D feature extractor, ShuffleNet V2\cite{ma2018shufflenet} to extract the semantic features. We preserve the temporal length, N by padding appropriately and reduce the spatial dimensions to 1 by performing global average pooling to the extracted frame features.

\subsection{Melspectogram Decoder}

The speaker embeddings are tiled  and concatenated to the frame features according to the frame sequence length N.

\begin{figure}
\includegraphics[width=\textwidth]{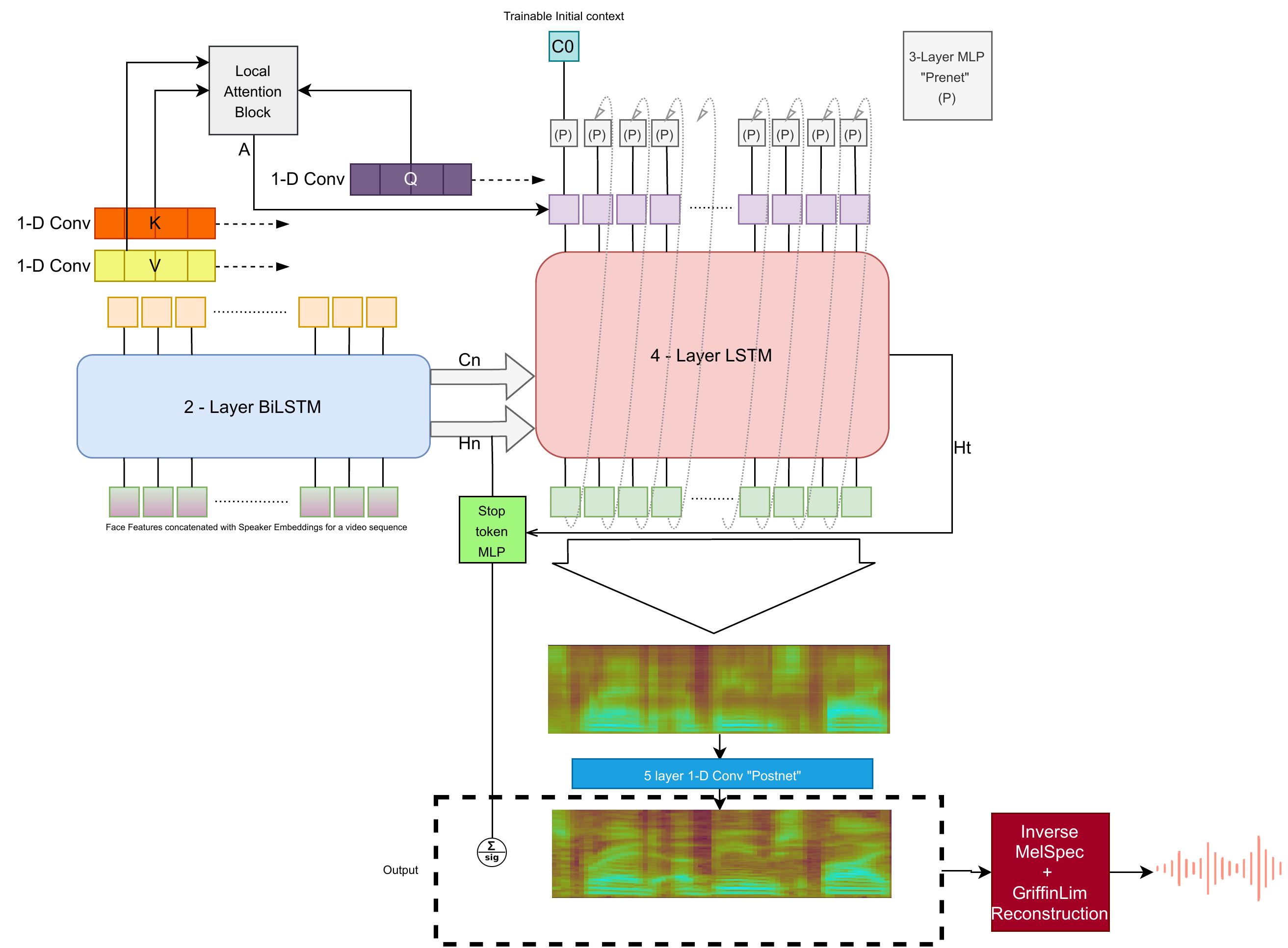}
\caption{The melspec decoder encodes the frame sequence into a condensed latent represetation and decoded melspec-frame auto-regressively. And attention mechanism is used to improve the temporal semantic flow.} \label{fig:deco}
\end{figure}

These "visual features" are taken in by the melspectogram decoder, as shown in \autoref{fig:deco} which decodes them to melspectogram frames(melspec-frame) in a auto-regressive manner. Since the frame sequence length N, and the number of time steps T does not need to be the same, we first encode the visual features into a latent representation using the 2-layer BiLSTM's hidden and cell state. We use a 4-layer LSTM\cite{hochreiter1997long} initialized with the BiLSTM's hidden and cell state to perform the decoding. The initial melspec-frame input to the decoder is trainable and jointly optimized along with the network. And, the regression of each time-step is done by passing the previous melspec-frame to an bottleneck network with heavy dropouts, called as "Prenet", which reduces the ground-truth bleeding during teacher forcing. It also forces the LSTM not to ignore the attended queries, as we restrict the information flow from the previous time-step. And to know when to stop the regression i.e., "stop token", we use a linear projection of the hidden state of each time-step concatenated with the hidden state of latent representation. The linear projection is activated using the sigmoid function, which acts as the gating value. During inference, we stop if the gating value is above a certain fixed threshold, with a default being 0.5.  

We use localized attention mechanism to improve the contextual information from the frame sequences, as the condensed latent encoding of the frame sequences will not be able to completely represent the temporal semantic flow. We generate the key, value by convolving 1-D convolution filters across the BiLSTM's outputs. And the query is generated for each time-step by using the previous melspec-frame and current hidden state. We add sinusoidal positional encodings\cite{vaswani2017attention} to key, query and value, to induce position-aware querying, thereby preventing previous queries to be concatenated for the query of the current time-step.

\section{Methodology}

\subsection{Datasets}

\subsubsection{AVSpeech}is a large-scale audio-visual dataset\cite{AVSpeech} . It consists of 3 to 15 second clips from YouTube videos.  We sample 33 thousand clips from this dataset and use it to train the Speaker Encoder. 

\subsubsection{Lip Reading in the Wild} consists of up to 1000 utterances of 500 different words, spoken by hundreds of different speakers\cite{Chung16}. We sampled 70 clips from the dataset for MOS and Face encoder evaluation. We also sample another 153 clips for Lip2Speech evaluation.

\subsubsection{UTKFace} \cite{zhang2017age} dataset is a large-scale face dataset with over 20,000 face images of people with ages ranging from 0 to 116 years old with different gender and ethnicity. We use UTKFace for training our face decoder. We randomly sample 3000 images from the dataset, then pass them to our speaker encoder to get embeddings and save them.

\subsection{Implementation Details}

\subsubsection{Training settings}

All experiments were implemented with PyTorch\cite{NEURIPS2019_9015}. We use the adam optimizer\cite{kingma2017adam} with initial learning rate set as 0.001 for training all experiments. We have developed a prototyping framework for models, where any change to the model, will generate appropriate model/validation/training logs automatically. We extensively use this to log our experiments.

To train the speaker encoder on AVSpeech, for every video sequence, a random frame is taken and we crop the face and its corresponding audio window, where the window length is chosen random from $1 \sim 3$ seconds. We train the network with a batch size of 64 and train until the contrastive criterion does not improve.     

The Lip2Speech network is trained for 300 epochs on both LRW and YLD, with early stopping when the validation accuracy does not improve for 10 epochs. We use a batch size of 84 and teacher force the decoder inputs. The teacher forcing is annealed for every 10 epochs, which slowly shifts the decoder to test sequence generation. We also augment the train data with random horizontal flips for the lip sequence. 

\subsubsection{Inference settings}

During inference time, the video frame sequence and a single frame of the sequence is given as the input to the network. We decode the melspec-frames auto-regressively until the stop-token threshold is greater than 0.5. After generating the melspectogram, we run Griffin-Lim Algorithm\cite{1172092} to reconstruct the raw audio waveform. And to reconstruct long sequences in the YLD dataset, we split the video into 2-second chunks and generate melspectogram for each chunk. We concatenate these chunks and apply Griffin-Lim to get the raw audio waveform. We found this method to be effective as the melspectogram inversion implicitly overlaps across melspec-frames to reconstruct the raw audio waveform, preventing abrupt discontinuities between adjoining audio chunks.

\section{Experiments}

\subsection{Speaker Encoder}

\subsubsection{Mean Opinion Score} The evaluation is done via the pre-trained Real-Time Voice Cloning repository which is based on SV2TTS\cite{jia2019transfer}. The network takes a text and a sample audio of a speaker to generate the given text in the given voice. The pipeline consists of three separately trained parts. First off a speaker encoder which takes an audio sample of the speaker as input and outputs a speaker embedding which the network will use to generate the correct voice. The next part is the synthesizer. Its input is the given text and the speaker encoding to output a melspectogram. The last part of the pipeline is a vocoder which generates raw audio from the melspectogram. The relevant design choice here is that the speaker encoder is separately trained and thus replaceable. We use this to give the model our speaker embeddings generated from the faces of people instead. With this, the pipeline produces an audio which we can compare to the ground truth and also to the audio generated by the original pipeline using an audio sample to encode the voice. 

For the evaluation, 70 samples from the Lip Reading in the Wild dataset\cite{Chung16} are passed through the pipeline. Once with the pre-trained speaker encoder and once with our speaker encoder. Then these 2 samples as well as the ground truth are graded with two metrics, their voice quality and the correlation between the voice and the image of the speaker. As shown in table\ref{MOS}, our model provides an overall better quality but the speaker audio embeddings produce voices with higher correlation. It is noticeable that the generated audio, no matter the given embeddings, is highly biased towards middle-aged white people. The network properly discerns between male and female with only a few outliners and different age groups have different amounts of energy to their voice. For both genders the voices do not deviate much from the mean voice. It is also perceptible that the voices generated for young students sound like adults. Also for different ethnicity's than white the generated audios sound distorted instead of producing a proper accent. 

\begin{table}
\caption{Voices generated with different embeddings and their respective MOS score compared to the ground truth.}
\label{MOS}
\begin{minipage}[t]{0.49\textwidth}
\begin{center}
    \begin{tabular}{|c|c|} 
    \hline
    \textbf{Voice} & \textbf{Quality}\\ 
  \hline
    ground truth & \textbf{4.56}\\
    speaker audio embedding & 3.37\\
    \textbf{speaker face embedding} & 3.55\\
 \hline
\end{tabular}
\end{center}
\end{minipage}
\begin{minipage}[t]{0.49\textwidth}
\begin{center}
\begin{tabular}{|c|c|} 
    \hline
    \textbf{Voice} & \textbf{Correlation}\\ 
  \hline
    ground truth & \textbf{4.44}\\
    speaker audio embedding & 3.12\\
    \textbf{speaker face embedding} & 3.03\\
 \hline
\end{tabular}
\end{center}
\end{minipage}
\end{table}

\subsubsection{Speaker Encoding Decoder}
In order to evaluate our speaker encoder qualitatively besides the voice quality, we need to visualise what the model has learned, i.e what traits of the face are captured in the speaker embedding. To achieve this goal, we train a face decoder that projects the 256 dimensional speaker embedding into a $64\times 64\times 3$ image. Our decoder is based on DCGAN\cite{radford2016unsupervised}. \\

\begin{figure}
\includegraphics[width=\textwidth]{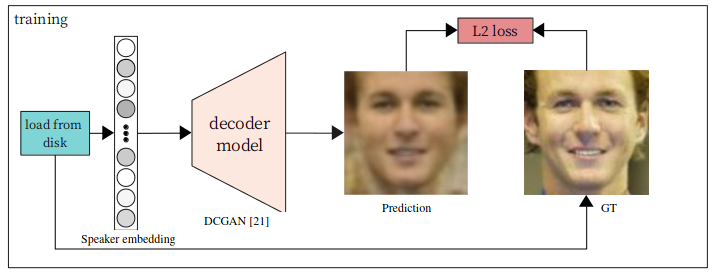}
\caption{Face decoder training process. Our decoder is based on \cite{radford2016unsupervised} and adapted to our face encoder's embedding output shape.}
\label{fig2}
\end{figure}

UTKFace dataset \cite{zhang2017age} was chosen to train the face decoder.  Then the decoder training process starts by passing the saved embeddings of 153 samples of the dataset and optimizing L2 distance between the reconstructed face and the GT as shown in Fig \ref{fig2}.

\subsubsection{Mean-Face Analysis} 
\begin{figure}
\includegraphics[width=\textwidth]{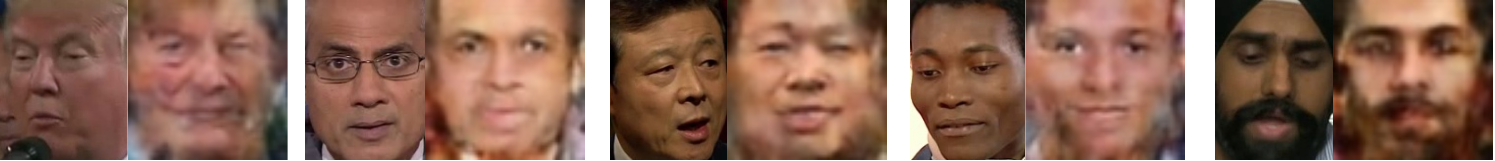}
\caption{Left is the ground truth and right is the reconstructed face.}
\label{MFA}
\end{figure}

The results of the decoder are visualized in Figure \ref{MFA}. As one can see the key features of the original face are reconstructed. The wrinkles of old people are still present which indicates that the voice of older people is successfully detected as different and thus encoded as a feature. The same can be said for gender features as well as skin color. Non-important features for the voice like glasses are removed. In terms of age we start to see some limitations, because children start to look like young adults. We also notice some unexpected results. For some reason some people smile while others do not. We were not able to link these smiles to any voice feature like a certain pitch and further testing would be required. Even more surprising is how beards are treated. One would assume that beards do not affect a persons voice and yet beards remain clearly visible. They often get reduced but never completely vanish. Related to this is the last image in Figure \ref{MFA} where the brown skin color was lost and replaced with white skin while the beard remained very visibly. Again we were not able to link a beard to a voice feature.

\clearpage
\subsection{Lip2Speech}
\begin{figure}
\begin{center}
    \includegraphics[width=0.8\textwidth]{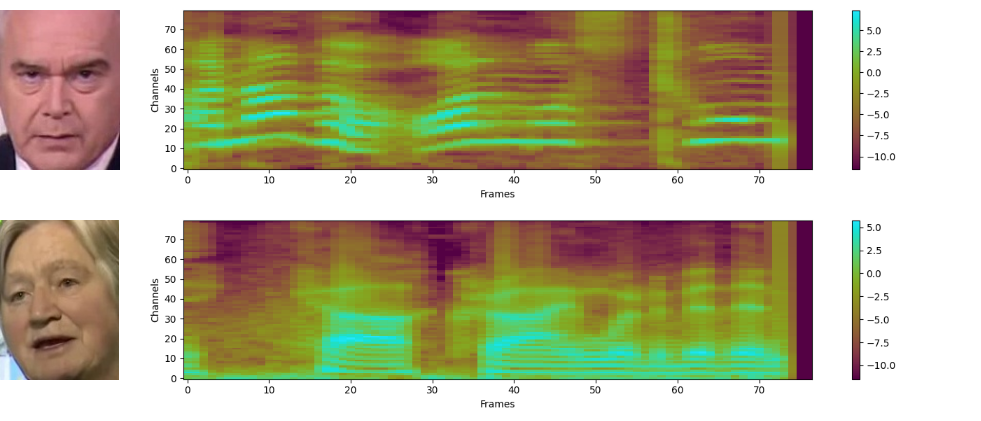}
\end{center}
\caption{Faces and the melspectograms generated with our Lip2Speech network.}
\label{melspec}
\end{figure}

We use LRW dataset for evaluating our model. We sample 153 videos from the whole dataset with people of different age, gender, ethnicity and accent. In fig \ref{melspec}, sample faces and the corresponding generated melspectograms are shown. We show Short-time Objective Intelligibility (STOI), extended STOI (ESTOI), Perceptual Evaluation of Speech Quality (PESQ), and Word Error Rate (WER) metrics results for the generated speech in table \ref{t2}. STOI, ESTOI, and PESQ metrics measure the speech intelligibility which is the the degree to which speech sounds (whether conversational or communication-system output) can be correctly identified and understood by listeners.  We also show results of Lip2Wav\cite{prajwal2020learning} and Chung et al.\cite{Chung_2017} in table \ref{t3}. The reason we add them in a different table is that it would be an unfair comparison since their results were obtained using LRW test split while we only use 153 random samples from the whole dataset. 

\begin{table}
\centering
\caption{Quantitative results of our model on the 153 test samples}
\begin{tabular}{|c|c|c|c|c|} 
    \hline
    & STOI \textuparrow & ESTOI \textuparrow & PESQ \textuparrow & WER \textdownarrow\\ 
  \hline
 \textbf{Lip2Speech} & 1.38 & 0.66 & 0.42 & 26.1\%  \\ 
 \hline
\end{tabular}
\label{t2}
\end{table}

\begin{table}
\centering
\caption{Quantitative results of other models on LRW test split}
\begin{tabular}{|c|c|c|c|c|} 
    \hline
    & STOI \textuparrow & ESTOI \textuparrow & PESQ \textuparrow & WER \textdownarrow\\ 
  \hline
 \textbf{Lip2Wav} & 0.543 & 0.344 & 1.197 & 34.3\%  \\ 
 \hline
  \textbf{Chung et al.} & NA & NA & NA & 38.8\%  \\ 
  \hline
\end{tabular}
\label{t3}
\end{table}

\clearpage
\subsection{YouTube Lip Data (YLD)}

The YouTube lip data(YLD) can transform any YouTube video into a Lip2Speech dataset. Given a YouTube video link and a target face that needs to be captured, the video is split into short segments of $1\sim3$  seconds. And only segments which contain the target face are taken by by performing face recognition\cite{bulat2017far} on every frame.

\begin{figure}
\includegraphics[width=\textwidth]{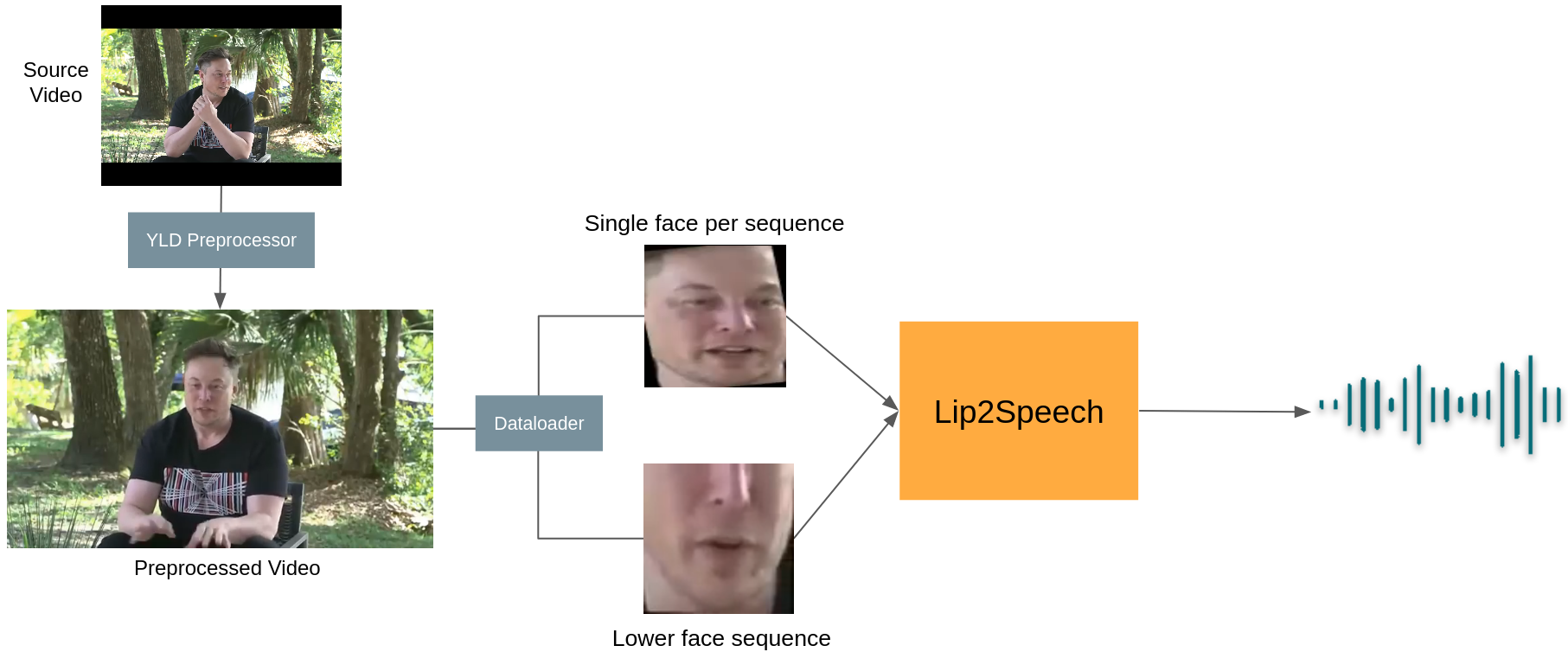}
\caption{The YLD pipeline can transform any Youtube video into training/testing data.} \label{fig:yld}
\end{figure}

\section{Limitations and Future Work}

First off we note that our subset of AVSpeech\cite{AVSpeech} to train the speaker encoder has been chosen randomly and thus might contain a bias towards a certain group of people. From the samples created for the MOS with our speaker face embeddings we observe that the generated voice is lacking in variation. Also the mean-face analysis shows that some features are not represented and need further learning. A bigger and more balanced training set should improve our results on this end.

Lip2Speech requires a lot of training data, and the generated voice sounds robotic. A way to alleviate the robotic voice is to use a neural vocoders\cite{oord2017parallel,oord2016wavenet} rather than optimization based Griffin-Lim Algorithm\cite{1172092}. And to improve the generalization and reduce the training data, the vocabulary of the words need to be clustered. This takes us in a direction where we also learn the visual and linguistic cues present in the context of lip reading.

\section{Concluding Remarks}

In this work, we proposed a method to synthesize speech from lip movements. Specifically, we focused on generating different vocal identity for individual speakers based on their age, gender, ethnicity. As our method uses the speaker embedding to incorporate speaker-specific information in the speech synthesis step, which allows the model to be used for all speakers, even ones that were previously unseen during training. We have evaluated
our model with extensive quantitative metrics and human studies. All the code and data for our work has been made publicly available$^4$.

\bibliography{bibliography}

\end{document}